\documentclass{INTERSPEECH2023}

\interspeechcameraready 

\usepackage{tablefootnote}

\title{Unsupervised Dialogue Topic Segmentation in Hyperdimensional Space}
\name{Seongmin Park$^1$, Jinkyu Seo$^2$, Jihwa Lee$^1$}
\address{
  $^1$ActionPower, Seoul, Republic of Korea\\
  $^2$Dept. of Applied Biology and Chemistry, Seoul National University, Seoul, Republic of Korea}
\email{seongmin.park@actionpower.kr, seojk413@snu.ac.kr, jihwa.lee@actionpower.kr}

\begin{document}

\maketitle
 
\begin{abstract}
We present \textit{HyperSeg}, a hyperdimensional computing (HDC) approach to unsupervised dialogue topic segmentation. HDC is a class of vector symbolic architectures that leverages the probabilistic orthogonality of randomly drawn vectors at extremely high dimensions  (typically over $10,000$). HDC generates rich token representations through its low-cost initialization of many unrelated vectors. This is especially beneficial in topic segmentation, which often operates as a resource-constrained pre-processing step for downstream transcript understanding tasks. HyperSeg outperforms the current state-of-the-art in 4 out of 5 segmentation benchmarks -- even when baselines are given partial access to the ground truth -- and is 10 times faster on average. We show that HyperSeg also improves downstream summarization accuracy. With HyperSeg, we demonstrate the viability of HDC in a major language task. We open-source HyperSeg to provide a strong baseline for unsupervised topic segmentation.\footnote{https://github.com/seongminp/hdseg}
\end{abstract}

\section{Introduction}

Automatic speech recognition (ASR) is often paired with downstream natural language processing (NLP) tasks for information extraction. Appropriately segmenting ASR transcripts to conform to the limited input length of downstream models is a practical and important consideration for such hybrid systems. Conventional NLP models cannot process whole meeting transcripts (which typically include hundreds or thousands of utterances spanning several hours) in a single inference context. Such physical constraints necessitate effective methods to divide transcripts into coherent segments before passing them on to downstream models.

Much research has focused on unsupervised approaches for topic segmentation due to the scarcity of labeled datasets with diverse domain and lexical coverage. Because segmentation is performed at the utterance level, dominant approaches to unsupervised topic segmentation use neural utterance embeddings. We find two problems with this line of research: neural utterance embeddings are very brittle to the domain in question \cite{wang2021x, wang-etal-2022-measure}, and existing unsupervised topic segmentation methods cannot determine segment counts on their own. Even state-of-the-art (SOTA) systems rely on carefully tuned hyperparameters, without which the systems offer lower segmentation accuracy than random segmentation. 

This work exploits hyperdimensional computing (HDC) to create a robust topic segmentation system resilient to domain and hyperparameter change. HDC leverages vector interactions in high-dimensional spaces to create scalable, aggregated representations. We initialize each token in an utterance as a random $10,000$-dimensional vector and bind those token representations to portray a complete utterance. HDC's aptitude for single-pass learning makes it inherently robust to domain shift \cite{10.1145/3524067}. Compared to neural approaches, HDC produces more representative utterance embeddings in a fraction of the processing time (Figure 1). We summarize our contributions as follows:
\begin{enumerate}
\item We introduce HyperSeg, the first framework to use HDC for topic segmentation.
\item HyperSeg outperforms the best unsupervised transcript segmentation baselines across multiple domains, even when baselines are provided with optimal hyperparameters and partial ground truth. 
\item HyperSeg is an order of magnitude faster than neural baselines, at an average of 10 times per utterance. HyperSeg achieves such speedup while running entirely on the CPU. 
\item HyperSeg increases downstream summarization accuracy compared to baseline and naive segmentation schemes.
\end{enumerate}

\begin{figure}[t]
\begin{minipage}{\columnwidth}                                                                                         
  \centering
  \includegraphics[width=\linewidth]{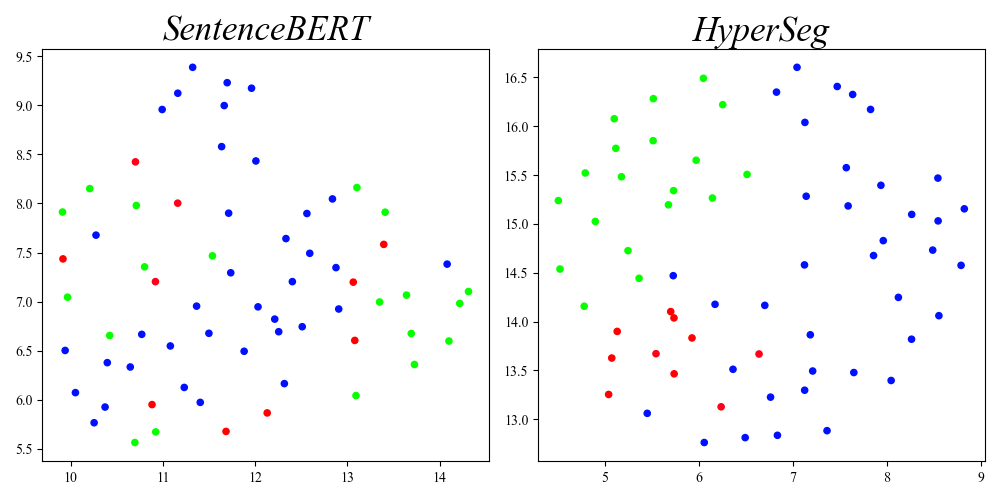}
  \caption{Utterance embeddings from Wiki-727k, created with SentenceBERT\protect\footnotemark(left) and HyperSeg (right). The color of each dot represents its topic. Representations were projected with t-SNE\cite{van2008visualizing}.}
  \label{fig:tsne}
\end{minipage}
\end{figure}

\footnotetext{https://huggingface.co/sentence-transformers/all-mpnet-base-v2. SentenceBERT \cite{reimers-gurevych-2019-sentence} performed better than more recent SimSCE \cite{gao-etal-2021-simcse}.}

\section{Related work}

\subsection{Hyperdimensional computing}

The concentration of measure \cite{ledoux2001concentration} is a phenomenon that occurs among a set of randomly initialized $n$-dimensional vectors where -- given sufficiently large $n$ -- the cosine similarity between each vector is heavily concentrated around $0$. Such a set of vectors are said to be pseudo-orthogonal to each other. In other words, a random vector drawn in high-enough dimensions will be probabilistically dissimilar to those previously drawn. Hyperdimensional computing (HDC) \cite{kanerva2009hyperdimensional} is a class of vector symbolic architectures that leverages such convenient orthogonality in high-dimensional space. HDC designates a random large dimensional vector (with $n >= 10,000$) to represent concepts of varying granularity, such as a single text token, a sentence, or an entity class. 
	
HDC occupies the middle ground between symbolic artificial intelligence (e.g. agent-based modeling) and distributional artificial intelligence (e.g. neural networks), amending the shortcomings of both. In symbolic architectures, one-hot entity representations render entity similarity calculations discontinuous and thus non-interpolable \cite{kleyko2023survey,marcus2020next}. At the opposite end of the spectrum, distributional semantics uses vector representations for natural interpolation and loss propagation, but cannot consolidate learned vectors into order- and dimension-preserving sequences (known as the “binding problem” \cite{greff2020binding}). HDC retains the benefits of distributed vector representations while offering a solution to the binding problem by tapping into the aforementioned pseudo-orthogonality of vectors at high dimensions. At extremely high vector spaces, the pseudo-orthogonality of randomly realized vectors converges to exact orthogonality. The architecture also comes with the benefit of extremely fast and parallelizable representation construction.

HDC has seen adoption in time-sensitive real-time systems \cite{kleyko2023survey}, often in conjunction with artificial neural networks. The framework has found its place in text processing as well \cite{najafabadi2016hyperdimensional}, boasting high accuracy and extremely fast processing time. For the first time, we apply HDC's representational power to transcript segmentation. We find that utterance embeddings created with HDC are semantically more coherent than ones created with the best sentence embedding neural networks.

\subsection{Unsupervised topic segmentation}

SOTA unsupervised topic segmentation algorithms first construct vector representations of each transcript utterance, to be delivered to downstream sequence segmentation algorithms. BERTSeg \cite{solbiati2021unsupervised} and CohereSeg \cite{xing-carenini-2021-improving} use SentenceBERT \cite{reimers-gurevych-2019-sentence} embeddings to represent each utterance. BERTSeg uses off-the-shelf pre-trained BERT \cite{devlin-etal-2019-bert} models, while CohereSeg further trains BERT with utterance-pair coherence loss. Semantic utterance vectors are fed to the TextTiling algorithm \cite{hearst1997text}, which is a linear sequence segmentation scheme that determines topic segments by vector distance. GraphSeg \cite{glavas-etal-2016-unsupervised} uses an alternative segmentation scheme by constructing a semantic graph with each utterance as a node. Utterance communities are determined in the semantic graph, and maximal cliques in the graph are designated as transcript segments.   

Depending on the transcript domain, however, the representational capability of semantic embeddings can falter. Each utterance embedding contains an unexplainable juxtaposition of semantic and lexical concepts \cite{wang2021x, pham2021out}. We find existing transcript segmentation algorithms are brittle in both representation construction and downstream segmentation. Such brittleness makes SOTA baselines extremely sensitive to hyperparameters, often resulting in lower accuracy than random segmentation. We show that replacing the utterance embedding step with HDC produces more topic-aware utterance vectors across multiple datasets. Leveraging such representational advantages, HyperSeg eliminates all hyperparameters while outperforming previous SOTA, even when optimal hyperparameters (including partial ground truth, such as the total segment count of a transcript) are provided to baselines. HyperSeg also operates an order of magnitude faster than neural baselines, which are often unfit for resource-constrained preprocessing tasks such as topic segmentation. 

\begin{figure}[t]
  \centering
  \includegraphics[width=\linewidth]{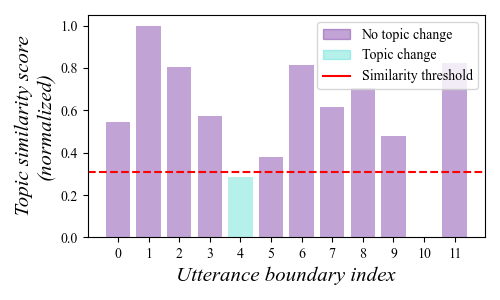}
  \caption{Normalized similarity scores from Doc2Dial. Utterance boundaries with similarity under an adaptively determined threshold are considered topic boundaries.}
  \label{fig:valley}
\end{figure}

\begin{table}[th]
  \caption{Summary of benchmark datasets. "Utts" and "Utt len" stand for mean utterance count and mean utterance length.}
  \label{tab:datasets}
  \centering
  \begin{tabular}{ccccc}
    \toprule
    \textbf{Name} & \textbf{Domain} & \textbf{Utts} & \textbf{Utt len} & \textbf{Segments}\\
    \midrule
    AMI & Meeting & $364.10$ & $36.88$ & $6.82$ \\
    ICSI & Meeting & $879.51$ & $50.84$ & $16.12$ \\ 
    Doc2Dial & Customer service & $30.13$ & $77.57$ & $1.42$ \\ 
    VT-SSum & Lecture & $306.10$ & $82.72$ & $33.87$ \\
    Wiki-727k & General & $46.43$ & $125.36$ & $4.66$ \\
    \bottomrule
  \end{tabular}
\end{table}

\section{Segmentation with HyperSeg}

\begin{table*}[th]
  \caption{Transcript segmentation results. HyperSeg performs best even without the ground-truth segment count provided to baselines. "BS" stands for boundary similarity. Results are averaged over 5 runs with different random seeds. Higher is better for both metrics. }
  \label{tab:segment_results}
  \centering
  \begin{tabular}{l||c|c||c|c||c|c||c|c||c|c}
    \toprule
    & \multicolumn{2}{c}{\textbf{AMI}} & \multicolumn{2}{c}{\textbf{ICSI}}  & \multicolumn{2}{c}{\textbf{Doc2Dial}}  & \multicolumn{2}{c}{\textbf{VT-SSum}} & \multicolumn{2}{c}{\textbf{Wiki-727k}} \\   
    \midrule
     \textbf{Model} & \textit{F1} & \textit{BS} & \textit{F1} & \textit{BS} & \textit{F1} & \textit{BS} & \textit{F1} & \textit{BS} & \textit{F1} & \textit{BS} \\
     \midrule
    Random & 5.19 & 2.75 & 5.09 & 2.64 & 10.76 & 6.89 & 23.91 & 13.30& 28.37 & 16.60 \\
    GraphSeg \cite{glavas-etal-2016-unsupervised} & 4.22 & 2.22 & 2.75 & 1.43 & 10.06 & 5.35 & 2.28 & 1.18 &21.09 & 12.26\\
    BERTSeg \cite{solbiati2021unsupervised} & 4.43 & 2.36 & 1.58 & 0.81 & 2.95 & 1.56 & 8.84 & 4.75 & 17.91 & 10.23\\
    CohereSeg \cite{xing-carenini-2021-improving} & 8.77 & 4.63 & \textbf{7.79} & \textbf{4.03} & 16.58 & 10.09 & 25.98 & 14.57 & 28.41 & \textbf{17.60}\\
    HyperSeg (ours) & \textbf{9.41} & \textbf{4.93} & 6.94 & 3.55 & \textbf{18.49} & \textbf{10.54} & \textbf{27.35} & \textbf{15.15} & \textbf{29.82} & 17.45\\
    \bottomrule
  \end{tabular}
\end{table*}

\subsection{Constructing utterance embeddings}
We follow \cite{najafabadi2016hyperdimensional} and create utterance embeddings by binding multiple hyperdimensional word embeddings. For every word in an transcript that is not a stopword, we randomly initialize a $10,000$-dimensional bipolar vector $\boldsymbol{w}$:
\begin{align}
\boldsymbol{w} = (x_0, …, x_{9,999}), \\
\text{where } x \sim \text{IID} \{-1, 1\}. \nonumber
\end{align}
Each element in the word embedding is either -1 or 1, and same words are mapped to the same embedding. High dimensionality ensures that each word embedding is roughly orthogonal to every other word embedding (“blessing of dimensionality” \cite{gorban2018blessing}). 

To represent a single utterance, we use permutation binding \cite{kleyko2023survey, najafabadi2016hyperdimensional} to consolidate all word embeddings in the utterance into a single hyperdimensional vector, also of dimension $10,000$. To generate an utterance vector $\boldsymbol{u}$, we first encode the position of each constituent word vector $\boldsymbol{w_{i}}$ by right-shifting ($\Pi$) $l-i-1$ times. $l$ is the utterance word count and $i$ is the zero-based word index within $\boldsymbol{u}$. The final utterance embedding is the component-wise majority vote ($M$) of all of its position-encoded word vectors ($\boldsymbol{w_i^{pos}}$): 
\begin{align}
    \boldsymbol{w_i^{pos}} = \Pi^{l - i - 1}(\boldsymbol{w_i}) \\
    \boldsymbol{u} = M(\boldsymbol{w_0^{pos}}, \boldsymbol{w_1^{pos}}, ...  \boldsymbol{w_{l-1}^{pos}})
\end{align} 
Topic segmentation does not require extracting constituent word embeddings after obtaining utterance embeddings. Even so, the reversibility of right shifts and position-wise mode selection guarantees the ability to recover each word embedding should the need arise.   

\subsection{Transcript segmentation}

We calculate the boundary similarity score ($S$) for every neighboring utterance pair $\boldsymbol{u_i}$ and $\boldsymbol{u_{i+1}}$: 

\begin{align}
    S = \frac{\boldsymbol{u_i} \cdot \boldsymbol{u_{i+1}}}{\lvert\boldsymbol{u_i}\rvert \lvert\boldsymbol{u_{i+1}}\rvert}
\end{align}

Each boundary score is the cosine similarity between surrounding utterance embeddings. An utterance boundary index $j$ has to satisfy two conditions to be selected as a topic boundary: its similarity score $s_j$ has to be a local minimum, and $s_j$ has to be lower than an automatically determined threshold $T$. 

Since the desired number of total segments is unknown, selecting a similarity threshold for topic boundaries is a non-trivial process. Baselines rely on hyperparameters to determine the total segment count, which often hinders segmentation accuracy at test time. HyperSeg automatically calculates the similarity cutoff threshold $T$ as one standard deviation of boundary scores ($\sigma_{s}$) subtracted from the mean ($\mu_{s}$). We design $T$ in such a way to incorporate both local (cosine distance between neighboring embeddings) and global (mean and standard deviation of all distances) statistics. The number of topics in a document is therefore adaptively determined.

Local minima boundary indices from $S$'s index set $I(S)$ with scores below $T$ are selected as the set of final segment boundaries $B$  (Figure \ref{fig:valley}):

\begin{align}
    T = \mu_{s} - \sigma_{s} \\
    B = \{j \in I(S) | s_j \, \text{is a local mimimum} \land s_j < T \}
\end{align}

HyperSeg can be optionally operated in \textit{damp mode}, where the number of segments is capped with a logarithm filter. Of utterance boundaries with similarity smaller than $T$, we select $N$ boundaries with the smallest boundary cosine similarity, where
\begin{align}
    N = \lfloor \log^2\lvert\{i \mid scores_i<T\}\rvert \rfloor
\end{align}
Dampened HyperSeg creates segments less eagerly.

\section{Experiments}

We measure HyperSeg’s effectiveness with two approaches: raw segmentation performance (accuracy and speed) and improvements in downstream summarization of segmented text. We also compare simple word- and n-gram counting with HyperSeg to empirically illustrate how HyperSeg performs beyond naive surface-level lexical matching. We use GraphSeg \cite{glavas-etal-2016-unsupervised}, BertSeg \cite{solbiati2021unsupervised}, and CohereSeg \cite{xing-carenini-2021-improving} as state-of-the-art baselines.

\subsection{Datasets and metrics}

Segmentation accuracy of HyperSeg and baselines are tested on five benchmarks: \textbf{AMI} \cite{kraaij2005ami}, \textbf{ICSI} \cite{janin2003icsi}, \textbf{Doc2Dial} \cite{feng2020doc2dial}, \textbf{VT-SSum} \cite{lv2021vt}, and \textbf{Wiki-727k} \cite{koshorek-etal-2018-text}. All datasets except Wiki-727k consist of dialogue transcripts. We include Wiki-272k to demonstrate that HyperSeg's robustness extends to written text segmentation. Selected benchmarks cover diverse domains, including meetings, customer service, lectures, and others. Table \ref{tab:datasets} includes detailed statistics of each dataset.

Linear segmentation is evaluated as a binary classification problem for each possible utterance boundary. Segmentation accuracy is measured with classification F1 scores and Boundary Similarity \cite{fournier-2013-evaluating}. The F1 measure captures the harmonic mean between precision and recall for the binary classification problem (boundary-or-not). Boundary Similarity measures the inter-coder agreement between system and reference segmentations.

Downstream summarization accuracy is measured by how much the system summary overlaps with the ground-truth summary. We measure ROUGE-1 (\textbf{R1}), ROUGE-2 (\textbf{R2}), and ROUGE-L (\textbf{RL}) \cite{lin-2004-rouge} between reference and system summaries. An off-the-shelf BART \cite{lewis-etal-2020-bart} model trained on the standard CNN/Daily Mail dataset \cite{nallapati2016abstractive} is used as the downstream summarizer.

\section{Results}

\begin{table*}[th]
  \caption{Average milliseconds elapsed per utterance. HyperSeg is 10 times faster than the next fastest baseline.}
  \label{tab:segmentation_time}
  \centering
  \begin{tabular}{l||c|c|c|c|c||c|c}
    \toprule
    \textbf{Model} & \textbf{AMI} & \textbf{ICSI} & \textbf{Doc2Dial} & \textbf{VT-SSum} & \textbf{Wiki-727k} & \textbf{Mean} & \textbf{$\times$ HyperSeg}\\
    \midrule
    GraphSeg \cite{glavas-etal-2016-unsupervised} & 11.60 & 13.71 & 10.66 & 11.11 & 9.73 & 11.36 & $\times$ 10.42\\
    BERTSeg \cite{solbiati2021unsupervised} & 11.92 & 12.02 & 12.82 & 11.45 & 11.22 & 11.89 & $\times$ 10.91 \\
    CohereSeg \cite{xing-carenini-2021-improving} & 15.99 & 14.75 & 15.15 & 15.28 & 15.76 & 15.39 & $\times$ 14.12 \\
    HyperSeg (ours) & 0.77 & 0.70 & 1.19 & 1.28 & 1.49 & 1.09 & $\times$ 1 \\
    \bottomrule
  \end{tabular}
\end{table*}

\subsection{Transcript topic segmentation}
HyperSeg outperforms baselines in all datasets except ICSI, even when the compared systems use the most optimal hyperparameters (Table \ref{tab:segment_results}). For GraphSeg, BertSeg, and CohereSeg, we report their best possible scores by providing the true number of ground-truth  segments as a hyperparameter. Only HyperSeg and CohereSeg consistently outperform random segmentation. Performance of all systems noticeably degrades in AMI and ICSI, where true segment boundaries constitute less than 2\% of all possible breakpoints. For such sparse datasets where metrics fail to take near-misses into account, we also report downstream summarization performance in the next section. All reported results are averages after 5 runs with different random seeds.

On average, HyperSeg is 10 times faster per utterance than the next quickest baseline GraphSeg (Table \ref{tab:segmentation_time}). Reported segmentation speeds are measured on an AMD EPYC 7742 CPU and four 80GB A100 GPUs. All baseline runs were conducted on GPUs, while HyperSeg ran entirely on the CPU. 

\subsection{Downstream summarization performance}
To demonstrate the effect of accurate topic segmentation on downstream summarization, we apply HyperSeg to AMI and ISCI before summarizing the transcripts. Compared to no segmentation, naive segmentation of equal segment length, and random segmentation, HyperSeg always improves summarization in terms of ROUGE scores (Table \ref{tab:summary_results}). 

\begin{table}[th]
  \caption{Downstream summarization accuracy. Only HyperSeg yields consistently better ROUGE scores than random or naive baselines.}
  \label{tab:summary_results}
  \centering
  \begin{tabular}{l||c|c|c||c|c|c}
    \toprule
    & \multicolumn{3}{c}{\textbf{AMI}} & \multicolumn{3}{c}{\textbf{ICSI}} \\   
    \midrule
     \textbf{Model} & \textit{R1} & \textit{R2} & \textit{RL} & \textit{R1} & \textit{R2} & \textit{RL} \\
     \midrule
    None & 0.19 & 0.03 & 0.12 & 0.17 & 0.03 & 0.13 \\
    Random & 0.27 & 0.05 & 0.13 & 0.16 & 0.02 & 0.08 \\
    Uniform & 0.13 & 0.01 & 0.08 & 0.22 & 0.03 & 0.12 \\
    GraphSeg \cite{glavas-etal-2016-unsupervised} & 0.22 & 0.04 & 0.13 & 0.23 & 0.04 & 0.12 \\
    BERTSeg \cite{solbiati2021unsupervised} & 0.22 & 0.04 & \textbf{0.14} & 0.21 & 0.04 & 0.11 \\
    CohereSeg \cite{xing-carenini-2021-improving} & 0.27 & \textbf{0.06} & 0.12 & 0.15 & 0.04 & 0.08 \\
    HyperSeg & \textbf{0.32} & \textbf{0.06} & \textbf{0.14} & \textbf{0.30} & \textbf{0.08} & \textbf{0.15} \\
    \bottomrule
  \end{tabular}
\end{table}

\subsection{Difference to simple lexical matching}

\begin{figure}[t]
  \centering
  \includegraphics[width=\linewidth]{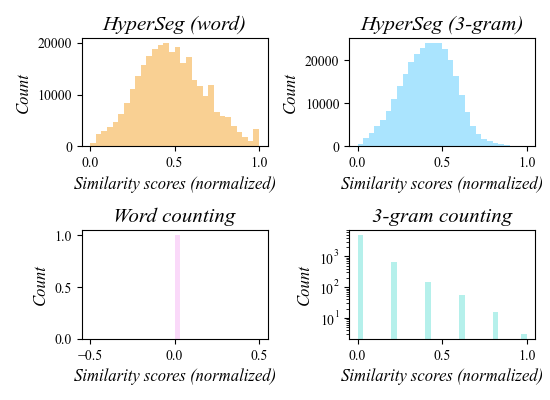}
  \caption{Distribution of utterance vector similarities (VT-SSum). Regardless of tokenization scheme (word-level or 3-gram), HyperSeg produces more gradual utterance similarity vectors compared to word or n-gram counting that use one-hot token vectors.}
  \label{fig:sims}
\end{figure}

HyperSeg offers two pronounced advantages over simple word or n-gram count-based topic segmentation: fine-grained sentence similarity scores and flexibility in token granularity. The distributed nature of HyperSeg produces a continuous spread of utterance boundary similarities (Figure \ref{fig:sims}). Word or n-gram counting yields a sparse spread of utterance similarities. Such segmentation renders the vectors unusable as features in possible downstream classification networks. HyperSeg can also accommodate character-level tokenization by modifying its binding algorithm. Such change in granularity offers a trade-off between robustness against character error rate and representational burden (an utterance vector has to contain more token vectors within a designated dimension compared to word-level tokenization). 

\section{Conclusion}

HyperSeg is a new state-of-the-art topic segmentation algorithm. We find that hyperdimensional computing's representational benefits can be effectively realized in dialogue topic segmentation. HyperSeg is more accurate and lightweight compared to best-performing baselines. We attribute HyperSeg's performance to the robustness of HDC embeddings across domains. Such advantages make HyperSeg an ideal component in ASR post-processing, which is often resource-constrained. We confirm that HyperSeg's representational capabilities extend to downstream summarization as well.

While this paper deals solely with linear transcript segmentation, a more natural approach to dialogue segmentation would take repetitive alternations of topics into consideration. Such methods require gathering related utterances into non-linear topical clusters. Equipped with the representational robustness of hyperdimensional vectors, HyperSeg can be naturally extended to accommodate such advanced schemes for segmentation. We also leave experiments with different utterance tokenization schemes (character-level, subword-level) for future research.
\bibliographystyle{IEEEtran}
\bibliography{mybib}

\end{document}